\documentclass[letterpaper, 10 pt, conference]{ieeeconf}
\IEEEoverridecommandlockouts
\usepackage{cite}
\usepackage{amsmath,amssymb,amsfonts}
\usepackage{algorithmic}
\usepackage{graphicx}
\usepackage{textcomp}
\usepackage[table]{xcolor}
\usepackage{booktabs}
\usepackage{tabularx}
\usepackage{multirow}
\usepackage{soul}
\usepackage{hyperref}

\setstcolor{red}

\newcommand{\IEEEtopfloatpad}{\vspace*{10pt}}

\newcolumntype{C}{>{\centering\arraybackslash}X}
\newcolumntype{Y}{>{\centering\arraybackslash}X}

\def\BibTeX{{\rm B\kern-.05em{\sc i\kern-.025em b}\kern-.08em
    T\kern-.1667em\lower.7ex\hbox{E}\kern-.125emX}}

\begin{document}

\title{A Causality-aware Infer-diagnose-refine Framework for Test-time Modality Adaptation in VLA Models}
\author{
Haoyu Zhang\textsuperscript{1},
Yuwei Wu\textsuperscript{1},
Jin Chen\textsuperscript{2,\dag},
Gao Zhi\textsuperscript{1,\dag},
Zhenxin Diao\textsuperscript{1},
Mingyang Gao\textsuperscript{1},
Kun Wu\textsuperscript{3},\\
Yongchun Liu\textsuperscript{2},
Fan Li\textsuperscript{2}\\[3pt]
\textsuperscript{1}Beijing Institute of Technology\\
\textsuperscript{2}AInnovation Co. Ltd.\\
\textsuperscript{3}Beijing Innovation Center of Humanoid Robotics\\[3pt]
\textsuperscript{\dag}Corresponding authors
}

\maketitle

\begin{abstract}

Vision-language-action (VLA) models predict sequential actions to execute tasks specified by language instructions, conditioned on visual observations and proprioceptive states. However, how to fuse modalities in VLA models remains an open problem, since robot manipulation involves dynamic phases, such as long-distance movements and close-range interactions, in which the importance of visual observations may vary over time. In this paper, we propose an infer-diagnose-refine (IDR) framework, a model-agnostic framework that can be integrated with diverse VLA architectures for refining action predictions at test time. IDR first infers actions under factual and counterfactual scenarios of visual observations, and then diagnoses the causal effects of visual observations as the estimated dynamic importance, which is finally used to refine the action predictions in a training-free manner. We further design a causality-aware action refiner to realize the IDR framework, including zero-padding interventions for inferring counterfactual actions, norm-based quantification for diagnosing causal effects, and gated residual fusion for refining actions. Extensive experiments on both simulation benchmarks and real-world tasks show improvements in overall performance across multiple VLA backbones, demonstrating the efficacy of dynamically adjusting visual importance at test time. The project page is at \href{https://floy4.github.io/IDR/}{here}.

\end{abstract}


\section{Introduction}
\label{sec:introduction}

Vision-language-action (VLA) models have emerged as a powerful paradigm for robot manipulation, learning to predict actions for a given language instruction based on visual observations and proprioceptive states~\cite{pi0,openvla,rt2,octo,pi05}.
Trained on large-scale robot datasets, VLA models exhibit good performance across diverse environments~\cite{xvla,vla_adapter}. 
However, the integration of multimodal information for action generation in VLA models is still an open problem.
Robot manipulation inherently involves dynamic processes, such as long-distance movements and close-range interactions, where visual observations may have varying importance, requiring a dynamic modality fusion framework during actual VLA deployment.

Recent studies have shown that visual information does not contribute uniformly to VLA models: the roles of vision and proprioception can vary across environments~\cite{fang2026vision,grant2026mechanistic} and, in particular, around motion-transition phases~\cite{lu2026fail}. These works estimate the visual importance and explore the dynamic utilization of visual information for action generation during training. This raises the following question: \textit{can visual importance be estimated and exploited for frozen VLA models at test time?} 
Such an approach avoids retraining, only incurring the inference-time computation.

In this paper, we explore a model-agnostic framework for adjusting the importance of visual observations in action generation at test time without retraining. 
This raises two key challenges: first, effectively estimating the dynamic visual importance for a VLA model; second, leveraging this estimation to refine action predictions during execution.
Our empirical analyses show that the visual importance varies not only across VLA execution stages, but also across model architectures and environments.
Different VLA architectures exhibit distinct visual information utilization patterns, ranging from vision-dominant to proprioception-dominant behavior.
Even for the same model, the role of visual observations can shift across environment settings.

\begin{figure}[t]
\centering
\includegraphics[width=\linewidth]{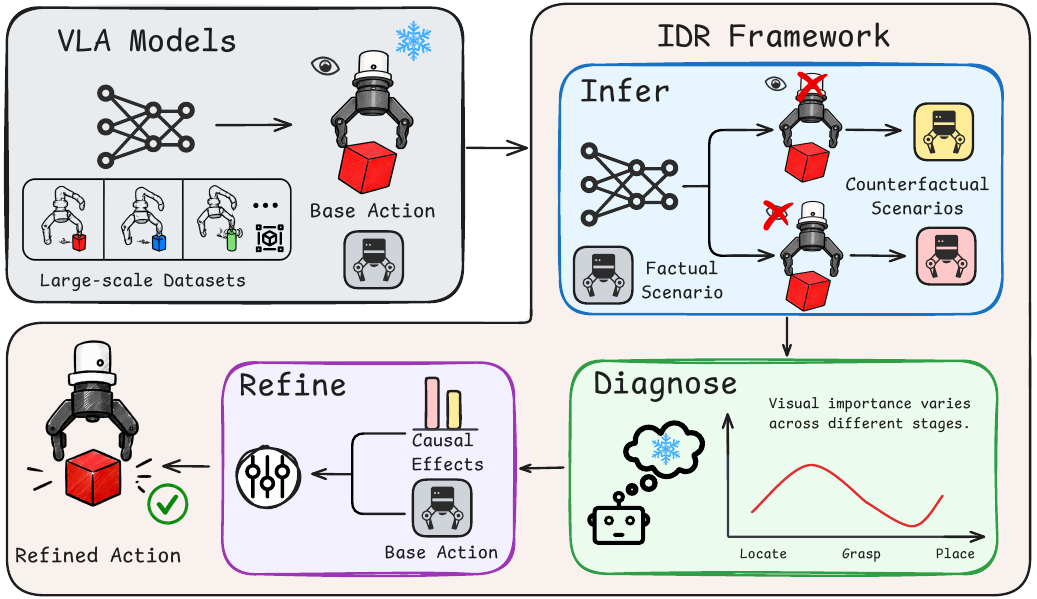}
\caption{Existing data-driven VLA models directly produce a base action from learned multimodal fusion. IDR infers factual and counterfactual outputs under modality interventions, diagnoses modality-specific causal effects, and refines the base action without retraining.}
\label{fig:teaser}
\vspace{-10pt}
\end{figure}

Building on these analyses, we propose \underline{i}nfer-\underline{d}iagnose-\underline{r}efine (IDR), a model-agnostic test-time framework that formulates visual importance through causal inference~\cite{pearl2009,peters2017}. Specifically, as shown in Fig.~\ref{fig:teaser}, at each timestep, IDR infers actions under factual and counterfactual scenarios, diagnoses the causal effects, and refines the base actions accordingly. Under this framework, we develop a causality-aware action refiner. 
It first constructs counterfactual inputs through zero-padding interventions, enabling the frozen VLA model to infer action predictions under modality interventions. 
The changes in these outputs are then quantified with a norm-based effect measurement to diagnose the dynamic visual importance. 
Finally, the diagnosed effects are selectively integrated with the base action prediction through gated residual fusion, producing a refined action while limiting excessive perturbations to low-level control.

Our contributions are summarized as follows:
\begin{itemize}
\item We propose IDR, a training-free model-agnostic framework for test-time modality adaptation. IDR follows three stages, infer-diagnose-refine, to estimate visual importance at each timestep and dynamically refine actions during task execution.

\item We design a causality-aware action refiner that estimates causal effects through counterfactual interventions and selectively applies gated residual fusion to refine action predictions.

\item We validate IDR across four VLA backbones, multiple simulation benchmarks, and real-world manipulation tasks, demonstrating that causality-aware modality adaptation at test time is a practical and broadly applicable strategy for improving VLA models.
\end{itemize}

\section{Related Work}
\label{sec:related}

\subsection{Vision-Language-Action Models}

Vision-language-action (VLA) models have rapidly evolved from task-specific robot policies toward general-purpose robot controllers built on large-scale vision-language and robot datasets~\cite{rt2,openvla,octo,pi0,pi05}.
Beyond scaling data and model capacities, some methods explore different designs in model architecture and action representation~\cite{xvla,vla_adapter}.
Despite these advances, the mechanisms shaping action generation during execution remain insufficiently understood.
A recent mechanistic study reveals that the visual pathway can strongly dominate action generation in VLA models, with language instructions frequently ignored when visual context alone determines the task~\cite{grant2026mechanistic}. Other studies examine modality imbalance through the lens of language grounding, showing that VLA models may produce visually plausible actions even when language instructions are insufficiently grounded~\cite{fang2026vision,zhang2026restoring}.
Another line examines phase-dependent failures in vision-proprioception policies, suggesting that proprioceptive signals may provide an easier-to-exploit predictive cue during motion-transition phases, causing policies to underuse visual observations~\cite{lu2026fail}.
Our work differs by targeting causal effects of visual observations at the action level. Rather than treating visual importance as a static property of the model, we estimate how visual observations causally affect the current action prediction and use it to guide action refinement during task execution.

\subsection{Test-Time Adaptation for VLA Models}

Recent work has explored test-time improvement for VLA models from different perspectives.
VLA-Pilot~\cite{vla_pilot} employs evolutionary diffusion to steer frozen VLA models at test time, achieving zero-shot deployment across embodiments.
Counterfactual comparison methods improve VLA policies by contrasting predictions under different input conditions.
CAG~\cite{fang2026vision}, for example, compares a language-conditioned policy with a language-unconditioned vision-action branch to mitigate vision-shortcut failures, whereas TAG~\cite{zhou2026tag} contrasts original and object-erased observations to provide residual guidance for instance-level grounding.
Although effective, these methods mainly target specific failures, including language-conditioning collapse and distractor-induced grounding errors.
Unlike these task-specific intervention strategies, IDR diagnoses the general causal effects of visual observations on the current action to adjust visual importance across manipulation stages without any retraining.

\subsection{Causal Inference in Deep Learning}

Causal inference~\cite{pearl2009,peters2017} provides a principled framework for reasoning about interventions and counterfactuals.
In machine learning, causal mediation analysis~\cite{vig2020} and causal abstraction~\cite{geiger2025} have been applied to understand the model's behavior beyond correlation.
In vision-language tasks, causal intervention
has addressed language bias in VQA~\cite{niu2021} and spurious correlations in scene graph generation~\cite{tang2020}.
Structured intervention pipelines such as simulate-analyze-reduce~\cite{chen2023metacausal} have shown that causal inference can analyze and reduce domain shifts for adaptation.
Our work extends this intervention-based perspective to VLA models, providing a framework for causality-aware test-time adaptation in robot manipulation.

\section{Preliminaries}
\label{sec:Preliminaries}

\begin{figure*}[t]
\IEEEtopfloatpad
\centering
    \includegraphics[width=\linewidth]{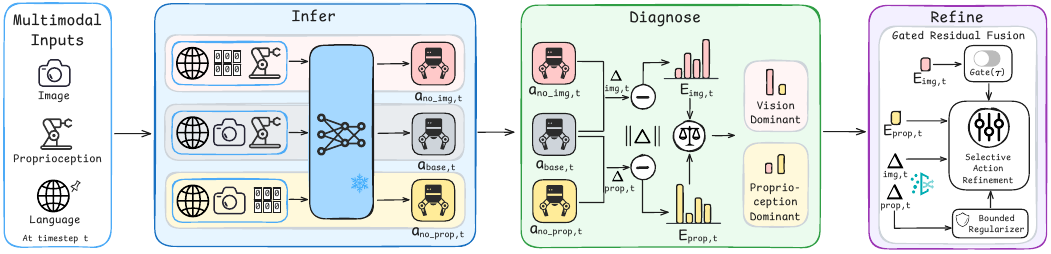}
    \caption{Overview of IDR framework. IDR keeps the VLA model frozen and refines its base action with causal effects estimated at test time. 
    The diagnosed effects guide a gated residual fusion, enabling training-free action refinement. The abbreviation ``prop'' denotes ``proprioception''.}
    \label{fig:method}
    \vspace{-10pt}
\end{figure*}

\subsection{VLA Model Formulation}
Given a task specified by language instructions, a vision-language-action (VLA) model predicts sequential robot actions conditioned on visual observations and proprioceptive states, serving as a crucial bridge between high-level human intentions and the physical world. 
Formally, let $V$, $S$, and $L$ denote the visual observation, proprioceptive state, and language instruction, respectively, and let $A$ denote the robot action. A VLA model $\pi_\theta (\cdot)$ learns to predict sequential actions from multimodal inputs, 
\begin{equation}
A = \pi_\theta \left(V, S, L\right),
\end{equation}
which models the conditional action distribution $p(A \mid V, S, L)$.
Concretely, given the visual observation $\mathit{v}_t$, proprioceptive state $\mathit{s}_t$, and instruction $\mathit{l}$ at timestep $t$, the action is predicted by 
\begin{equation}
\mathit{a}_t = \pi_\theta \left(\mathit{v}_t, \mathit{s}_t, \mathit{l} \right).
\end{equation}
The parameters $\theta$ of the VLA model are optimized on a dataset $\mathcal{D}$ by minimizing the action prediction loss:
\begin{equation}
\theta^{*} =\arg \min_{\theta} \mathbb{E}_{(\mathit{v}_t, \mathit{s}_t, \mathit{l}, \mathit{a}_t)\sim \mathcal{D}} \left[ \mathcal{L}_{\mathrm{act}} \left( \pi_\theta(\mathit{v}_t, \mathit{s}_t, \mathit{l}), \mathit{a}_t \right) \right],
\end{equation}
where $\mathcal{L}_{\mathrm{act}}$ denotes the action prediction loss determined by the VLA architecture, such as conditional flow-matching loss~\cite{pi0}, or autoregressive cross-entropy loss~\cite{rt2,openvla}.

\subsection{Causal Inference}
Causal inference~\cite{pearl2009,peters2017} provides a principled framework for reasoning about how input factors affect system outputs through interventions and counterfactual reasoning. We denote random variables as capital letters (\emph{i.e.,} $X,Y$) and their observed values as lowercase letters (\emph{i.e.,} $x,y$).

\textbf{Causal formulation.}
Consider a system governed by $K$ input factors $X_1, \dots, X_K$, a contextual variable $C$, and an output $Y$. Given observed values $X_1=x_1, \dots, X_K=x_K$ and $C=c$, the factual output distribution is
\begin{equation}
p(Y \mid X_1=x_1, \dots, X_K=x_K, C=c).
\end{equation}
The context $C$ remains fixed and defines the condition under which the causal effects of the input factors are studied.

\textbf{Counterfactual intervention.}
To isolate the causal effect of a specific factor $X_k$, we construct a counterfactual scenario by intervening on $X_k$, \emph{i.e.,} replacing its observed value $x_k$ with an alternative value $x_k^{*}$ while keeping all other factors $X_{-k}=\{X_j\}_{j \neq k}$ and context $C$ unchanged. Let $do(X_k=x_k^{*})$ denote this intervention and $Y_{x_k^{*}}$ denote the outputs under the counterfactual scenario. The distribution is formulated as
\begin{equation}
\label{eq:cf_dist}
p(Y_{x_k^{*}} \mid do(X_k=x_k^{*}), X_{-k}=x_{-k}, C=c).
\end{equation}

\textbf{Causal effect.}
The causal effect of $X_k$ on $Y$ under context $C$ is estimated by comparing the factual outputs $p(Y \mid X_1,\dots,X_k,C)$ with counterfactual outputs $p(Y_{x_k^{*}} \mid X_1,\dots,do(X_k=x_k^{*}),C)$. By repeating interventions for all input factors $k\in \{1,\dots,K\}$, their individual causal effects can be estimated, enabling a principled comparison of their relative contributions to the system's behavior.

\section{Method}
\label{sec:method}

\subsection{Infer-diagnose-refine Framework}
\label{sec:framework}
The infer-diagnose-refine (IDR) framework is motivated by the fact that VLA models exhibit dynamic reliance on visual information throughout task execution.
IDR conceptualizes the dynamic importance of visual observations as their causal effects on the outputs. The core intuition is to elicit a counterfactual reasoning process within a VLA model, imagining a scenario by omitting visual observations and asking: \emph{how would the predicted action change if the current visual observation were absent?} By systematically comparing this counterfactual prediction with the factual one, the causal effects of the visual observations can be explicitly diagnosed. 
IDR is illustrated in Fig.~\ref{fig:method}, which is a three-stage, training-free pipeline.

\textbf{Infer.} We construct counterfactual scenarios by applying interventions to the visual observation $V$ and the proprioceptive state $S$, respectively. 
By inferring the corresponding counterfactual outputs, we decouple the multimodal inputs 
for subsequent modality-specific causal diagnosis.

\textbf{Diagnose.} We quantify the dynamic importance of the visual observations by estimating their causal effects on the outputs. This is achieved by measuring the distributional shift between the factual and counterfactual outputs. The resulting magnitude serves as a diagnostic signal, explicitly reflecting the visual importance at the current timestep, and is further used to guide the action refinement.

\textbf{Refine.} Based on the diagnosed causal effects, we adjust the visual importance to dynamically recalibrate the action predictions. A \textit{gated residual fusion} scheme is designed where the counterfactual residual deviation is selectively weighted by the magnitude of the causal effect.

\subsection{Causality-Aware Action Refiner}
\label{sec:action_refiner}
Under the IDR framework, we design a causality-aware action refiner that translates the diagnosed causal effects into a gated residual fusion for action refinement. Crucially, this refiner dynamically adjusts visual importance while preserving the original behavior prior of the VLA model. 
Specifically, it sequentially 
instantiates the three stages of IDR with \emph{zero-padding interventions} for counterfactual inference, \emph{norm-based quantification} for modality-specific causal diagnosis, and \emph{gated residual fusion} for action refinement.

\begin{figure}[t]
\IEEEtopfloatpad
    \centering
    \includegraphics[width=\columnwidth]{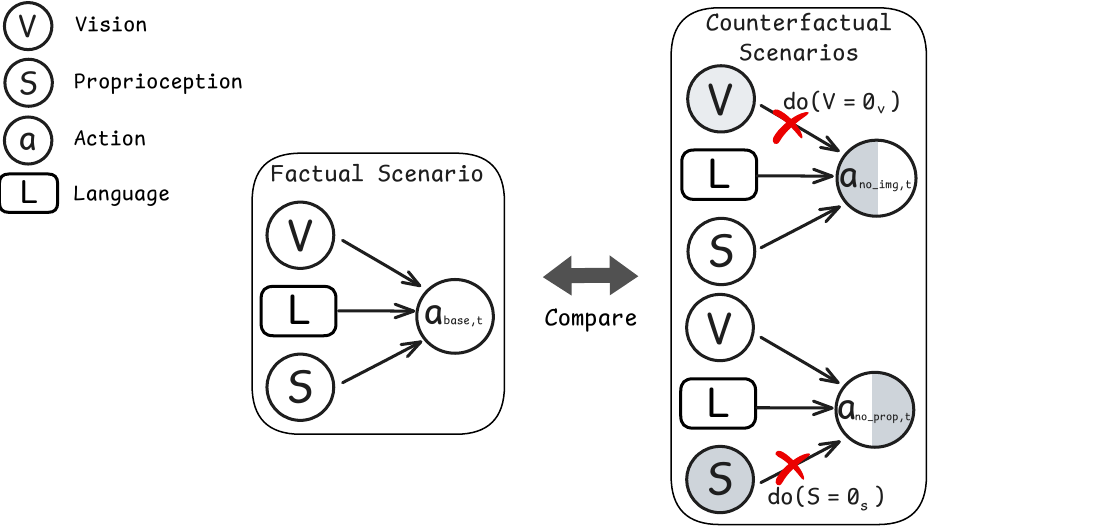}
    \caption{Causal graph of IDR. Causal effects are estimated by comparing the factual prediction with counterfactual ones under interventions.}
    \label{fig:causal_graph}
    \vspace{-10pt}
\end{figure}

We formulate the dynamic importance of visual observations as their causal effects on action prediction. We present a causal inference paradigm for VLA models by defining two input factors: the visual observation $X_1=V$ and the proprioceptive state $X_2=S$. The left side of Fig.~\ref{fig:causal_graph} represents the conditional distribution $p(Y \mid X_1, X_2, C)$, which corresponds to the conditional distribution $p(A \mid V, S, L)$ in VLA, and the language instruction $L$ serves as the context $C$. 
This formulation enables us to analyze the causal effects of visual and proprioceptive inputs on action prediction under a given language instruction, laying the theoretical foundation for the subsequent realization of the three stages in the causality-aware action refiner.

\emph{1) Infer: Zero-padding Interventions}
\label{sec:infer}

Given a VLA model $\pi_{\theta^{*}}$ with the frozen parameters $\theta^{*}$, we first compute the factual base action using the unmodified inputs.
We then obtain two counterfactual actions by applying zero-padding interventions to the visual and proprioceptive inputs, respectively, isolating the effects of each modality on the VLA model's predictions.

Specifically, given the multimodal inputs $(\mathit{v}_t, \mathit{s}_t, \mathit{l})$, the factual base action $\mathit{a}_{\text{base},t}$ at timestep $t$ is computed as
\begin{equation}
\label{eq:baseline}
\mathit{a}_{\text{base},t} =\pi_{\theta^{*}}
\left(\mathit{v}_t, \mathit{s}_t, \mathit{l} \right).
\end{equation}

We then use \(do(V=\mathit{0}_v)\) and \(do(S=\mathit{0}_s)\) to replace the respective input with an all-zero tensor of the same dimension. The corresponding counterfactual actions are computed as
\begin{equation}
\begin{aligned}
\mathit{a}_{\text{no\_img},t}
&= \pi_{\theta^{*}}\!\left(do(V=\mathit{0}_v),\mathit{s}_t,\mathit{l}\right)
=
\pi_{\theta^{*}}\!\left(\mathit{0}_v,\mathit{s}_t,\mathit{l}\right),\\
\mathit{a}_{\text{no\_prop},t}
&=\pi_{\theta^{*}}\!\left(\mathit{v}_t,do(S=\mathit{0}_s),\mathit{l}\right)
=\pi_{\theta^{*}}\!\left(\mathit{v}_t,\mathit{0}_s,\mathit{l}\right).
\end{aligned}
\end{equation}

These counterfactual actions represent the model's outputs when a specific modality is masked,
providing the basis for the dynamic importance estimation in the next stage.

\emph{2) Diagnose: Norm-based Quantification}
\label{sec:diagnose}

Building on the factual and counterfactual predictions from the infer stage, we now quantify the dynamic, time-varying importance of the visual modality by estimating its causal effect on the action predictions. This is achieved by measuring the distributional shift between the factual action and each counterfactual output, which explicitly reflects the modality-specific effects at the current timestep.

We first compare action predictions under the factual and counterfactual scenarios. The action deviations capture the directional causal effect of each modality since only the intervened modality is changed, computed as
\begin{equation}
\begin{aligned}
\Delta_{\text{img},t} &= \mathit{a}_{\text{base},t} - \mathit{a}_{\text{no\_img},t}, \\
\Delta_{\text{prop},t} &= \mathit{a}_{\text{base},t} - \mathit{a}_{\text{no\_prop},t}.
\end{aligned}
\end{equation}

We then quantify the magnitude of each modality's causal effect by computing the $L_2$ norm of the two deviation vectors
\begin{equation}
E_{\text{img},t} = \|\Delta_{\text{img},t}\|_2, \quad
E_{\text{prop},t} = \|\Delta_{\text{prop},t}\|_2.
\end{equation}
A larger $E_{\text{img},t}$ indicates a more significant distributional shift upon visual ablation, implying that the model heavily relies on visual inputs at timestep $t$ (\emph{e.g.}, during close-range interactions). Conversely, a small $E_{\text{img},t}$ suggests that the visual importance is suppressed (\emph{e.g.}, during long-distance movements).
The quantified magnitudes serve as training-free, adaptive diagnostic signals that estimate the dynamic importance of visual observations and will directly trigger the gated refinement mechanism in the subsequent stage.

\emph{3) Refine: Gated Residual Fusion}
\label{sec:refine}

In the refinement stage, our core objective is to preserve the VLA prior while adaptively regulating the importance of visual observations during dynamic action execution, thereby avoiding \emph{blind execution}—\emph{i.e.,} relying predominantly or entirely on proprioceptive states and ignoring visual observations. The tendency of the policy to act blindly can be intuitively assessed in an unsupervised manner based on the magnitude of the visual causal effect. Hence, we propose a Gated Residual Fusion mechanism to enable action refinement during blind execution, achieving a principled trade-off between preserving the VLA prior and dynamically adjusting the importance of visual observations.


Specifically, we employ a dynamic gating mechanism, which regulates the action refinement based on the magnitude of the visual causal effect $E_{\text{img},t}$. We define the gate as
\begin{equation}
g_t = \mathbb{I}\left[E_{\text{img},t} < \tau\right],
\end{equation}
where $\tau$ is the intervention threshold (set to the mean causal effect of visual observations under the baseline model), and $\mathbb{I}[\cdot]$ is an indicator function. When $E_{\text{img},t}$ falls below $\tau$, it indicates that the VLA model underutilizes visual cues at the current timestep (\emph{e.g.}, during long-distance movements where visual cues might be suppressed). In this case, the gate is activated ($g_t = 1$), and the refiner injects a correction $\Delta_{\text{img},t}$ to explicitly compensate for the suppressed visual importance. Conversely, when $E_{\text{img},t} \ge \tau$, the model already exhibits strong visual reliance (\emph{e.g.}, during close-range interactions), and the gate remains inactive ($g_t = 0$) to preserve the base action.

Directly applying the proprioceptive correction could induce high-frequency control jitter.
To limit such perturbations, we further introduce a bounded proprioceptive regularizer that scales and clips the residual as a stabilizing term,
\begin{align}
w_{\text{prop},t} &= \beta \cdot \min\!\left(1, \frac{E_{\text{prop},t}}{E_{\text{img},t} + \epsilon}\right), \\
u_{\text{prop},t} &= \operatorname{clip}\left(\Delta_{\text{prop},t}, -\lambda, \lambda\right),
\end{align}
where $w_{\text{prop},t}$ dynamically scales the proprioceptive correction relative to the diagnosed visual importance, $\beta$ controls the overall regularization strength, and $u_{\text{prop},t}$ is the clipped proprioceptive residual, which bounds $\Delta_{\text{prop},t}$ within $[-\lambda,\lambda]$ for smooth actuation. In all experiments, we set $\beta = 0.05$ and $\lambda = 0.1$.
The final refined action is formulated as
\begin{equation}\label{eq:final_action}
\mathit{a}_{\text{final},t} =
\mathit{a}_{\text{base},t} +
g_t \cdot \!\left(
\alpha \Delta_{\text{img},t}
+
w_{\text{prop},t} \cdot u_{\text{prop},t}
\right),
\end{equation}
where $\alpha$ controls visual refinement strength (set to $0.08$ in experiments). In this formulation, the diagnosed magnitude $E_{\text{img},t}$ acts as the adaptive trigger, the causal effect $\Delta_{\text{img},t}$ provides the targeted visual refinement, and the bounded proprioceptive term limits low-level control perturbations. Ultimately, this gated residual fusion realizes the test-time dynamic adjustment of visual importance.

\section{Experiments}
\label{sec:experiments}

\subsection{Simulation Experiments}

\textbf{Experimental setup.} We evaluate four VLA backbones with different architectures: $\pi_{0.5}$~\cite{pi05}, X-VLA~\cite{xvla}, VLA-Adapter~\cite{vla_adapter}, and OpenVLA-OFT~\cite{openvla_oft}.
Simulation benchmarks include LIBERO~\cite{libero}, SIMPLER~\cite{li24simpler}, and CALVIN ($ABC \rightarrow D$)~\cite{calvin}. We directly test publicly available weights under the same official evaluation protocols.

\textbf{Observations from test-time causal diagnosis.}
To analyze modality-specific impacts on model behavior, we first perform intervention-based diagnostics across evaluated models and environments. To enable fair comparisons, we average causal effects over all timesteps, obtaining the normalized visual importance ratio $R_{\text{img}} = E_{\text{img}} \,/\, (E_{\text{img}} + E_{\text{prop}} + \epsilon)$, where $\epsilon$ is a small constant for numerical stability. Based on this metric, we derive three empirical observations.

\begin{figure}[t]
\IEEEtopfloatpad
    \centering
    \includegraphics[width=\columnwidth]{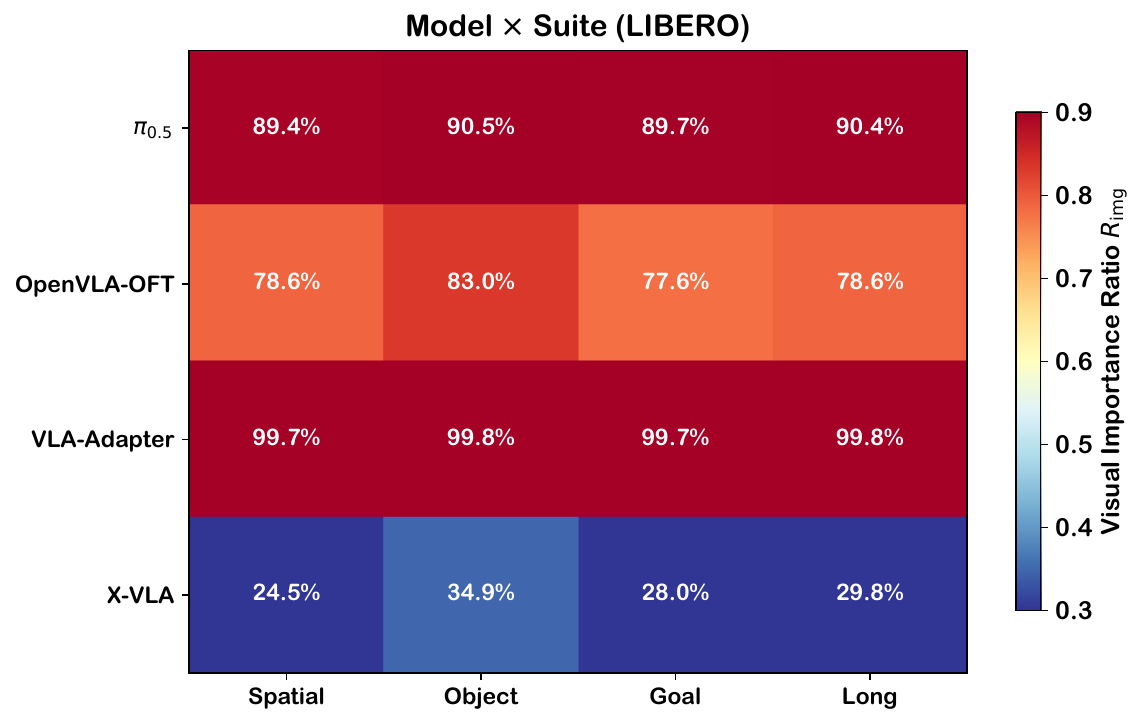}
    \caption{$R_{\text{img}}$ heatmap across VLA backbones and LIBERO suites. Each model maintains a distinct visual importance pattern.}
    \label{fig:baseline_suite_heatmap}
    \vspace{-10pt}
\end{figure}

\begin{table}[t]
    \centering
    \caption{X-VLA visual importance patterns across environments.}
    \label{tab:xvla_cross_env}
    \resizebox{\columnwidth}{!}{%
    \begin{tabular}{lcccc}
    \toprule
    \textbf{Environment} & \textbf{Suite/Task} & $E_{\text{img}}$ & $E_{\text{prop}}$ & $R_{\text{img}}$ \\
    \midrule
    LIBERO & All suites & 0.67 & 1.59 & 29.6\% \\
    SIMPLER & WidowX & 0.62 & 0.76 & 44.9\% \\
    CALVIN & ABC$\to$D & 0.61 & 0.60 & 50.4\% \\
    SIMPLER & Google Variant Aggregation & 1.18 & 0.09 & 92.9\% \\
    SIMPLER & Google Visual Matching & 0.99 & 0.07 & 93.4\% \\
    \bottomrule
    \end{tabular}%
    }
    \vspace{-10pt}
\end{table}

\begin{table}[t]
\centering
\footnotesize
\caption{LIBERO benchmark results (\%). Results marked with $^*$ are reproduced by us and \textbf{bold} indicates the best performance.}
\label{tab:libero_results}
\resizebox{\columnwidth}{!}{%
    \begin{tabular}{@{}l l cccc c@{}}
    \toprule
    Size & Method & Spatial & Object & Goal & Long & Avg \\
    \midrule
    \multirow{11}{*}{\textbf{\shortstack{Large\\($\geq$4B)}}}
     & UniVLA~\cite{univla} & \textbf{96.50} & 96.80 & 95.60 & 92.00 & 95.20 \\
     & UnifiedVLA~\cite{unifiedvla} & 95.40 & 98.80 & 93.60 & 94.00 & 95.50 \\
     & OpenVLA~\cite{openvla} & 84.70 & 88.40 & 79.20 & 53.70 & 76.50 \\
     & DD-VLA~\cite{ddvla} & 97.20 & 98.60 & 97.40 & 92.00 & 96.30 \\
     & PD-VLA~\cite{pdvla} & 95.50 & 96.70 & 94.90 & 91.70 & 94.70 \\
     & MolmoAct~\cite{molmoact} & 87.00 & 95.40 & 87.60 & 77.20 & 86.80 \\
     & ThinkAct~\cite{thinkact} & 88.30 & 91.40 & 87.10 & 70.90 & 84.43 \\
     & CoT-VLA~\cite{cotvla} & 87.50 & 91.60 & 87.60 & 69.00 & 83.93 \\
     & WorldVLA~\cite{worldvla} & 87.60 & 96.20 & 83.40 & 60.00 & 81.80 \\
    \cmidrule{2-7}
     & OpenVLA-OFT~\cite{openvla_oft}$^*$ & 92.60 & \textbf{99.20} & 96.20 & 94.60 & 95.65 \\
    \rowcolor{gray!15} & +IDR {\small(+0.90)} & 94.20 & \textbf{99.20} & \textbf{98.00} & \textbf{94.80} & \textbf{96.55} \\
    \midrule
     & $\pi_0$~\cite{pi0} & 96.80 & 98.80 & 95.80 & 85.20 & 94.20 \\
     & SmolVLA~\cite{smolvla} & 93.00 & 94.00 & 91.00 & 77.00 & 88.75 \\
     & GR00T-N1~\cite{grootn1} & 94.40 & 97.60 & 93.00 & 90.60 & 93.90 \\
    \cmidrule{2-7}
    \multirow{-2}{*}{\textbf{\shortstack{Small\\($<$4B)}}}
     & $\pi_{0.5}$~\cite{pi05}$^*$ & 97.50 & 95.00 & \textbf{97.00} & \textbf{95.50} & 96.25 \\
    \rowcolor{gray!15} & +IDR {\small(+1.25)} & \textbf{98.00} & \textbf{100.00} & \textbf{97.00} & 95.00 & \textbf{97.50} \\
    \midrule
    \multirow{7}{*}{\textbf{\shortstack{Tiny\\($<$1B)}}}
     & GraspVLA~\cite{graspvla} & -- & 94.10 & 91.20 & 82.00 & 89.10 \\
     & VLA-OS~\cite{vlaos} & 87.00 & 96.50 & 92.70 & 66.00 & 85.55 \\
     & UniAct~\cite{uniact} & 77.00 & 87.00 & 77.00 & 70.00 & 77.75 \\
    \cmidrule{2-7}
     & X-VLA~\cite{xvla}$^*$ & 92.60 & 99.20 & \textbf{98.00} & 94.40 & 96.05 \\
    \rowcolor{gray!15} & +IDR {\small(+0.45)} & \textbf{97.00} & \textbf{100.00} & 95.00 & 94.00 & \textbf{96.50} \\
     & VLA-Adapter~\cite{vla_adapter}$^*$ & 94.20 & 95.60 & \textbf{98.00} & 92.60 & 95.10 \\
    \rowcolor{gray!15} & +IDR {\small(+1.40)} & \textbf{97.00} & 98.00 & 96.00 & \textbf{95.00} & \textbf{96.50} \\
    \bottomrule
    \end{tabular}}
\end{table}

\textbf{Observation 1: Environments shift visual importance patterns.}
The same VLA model can shift its visual importance pattern substantially across different benchmarks, as shown in Table~\ref{tab:xvla_cross_env}. X-VLA is proprioception-dominant in LIBERO but becomes vision-dominant in the Google SIMPLER settings.
This suggests that visual importance patterns are associated with deployment environments.

\textbf{Observation 2: Model architectures shape distinct visual importance patterns.}
Within the same benchmark, different VLA models exhibit substantially different visual importance patterns. As shown in Fig.~\ref{fig:baseline_suite_heatmap}, models fall into distinct modality importance patterns: $\pi_{0.5}$, VLA-Adapter, and OpenVLA-OFT are vision-dominant while X-VLA is proprioception-dominant.
This variation may be associated with architectural designs where models relying heavily on vision-language representations maintain a higher $R_{\text{img}}$, while architectures that inject proprioceptive states and action tokens earlier show stronger proprioceptive effects.

\begin{figure}[t]
    \centering
    \includegraphics[width=\columnwidth]{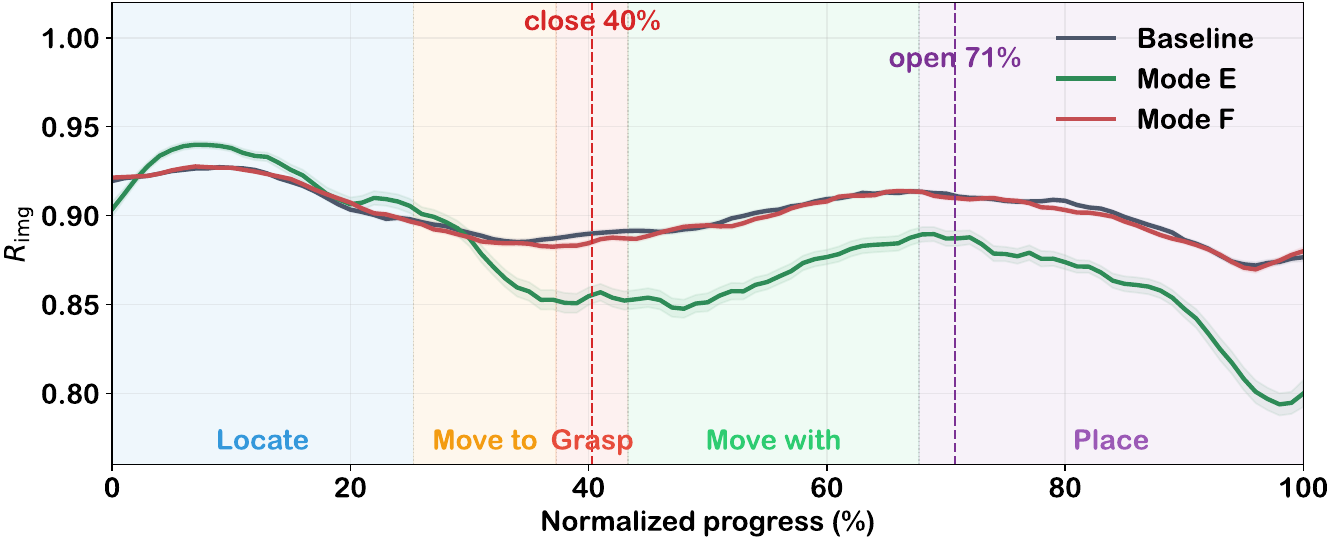}
    \caption{
    Phase-aligned visual importance ratio $R_{\text{img}}$ of $\pi_{0.5}$ in LIBERO.
    The plot reports $R_{\text{img}}$ over task progress for the baseline, Mode~E, and Mode~F.
    Dashed vertical lines indicate gripper closing and opening events.
    }
\label{fig:phase_aligned_effect_ratio}
\vspace{-10pt}
\end{figure}

\textbf{Observation 3: Manipulation phases alter visual importance patterns.}
Within a task execution, visual importance patterns change with manipulation phases.
As shown in Fig.~\ref{fig:phase_aligned_effect_ratio}, the baseline $R_{\text{img}}$ fluctuates with task progress and changes around gripper closing and opening events, highlighting the need for dynamic adaptation during execution.

These observations reveal that visual importance patterns are structured but not fixed---they depend on architectures, environments, and task phases. This insight supports our causality-aware action refinement: test-time intervention should be guided by the model's current causal effects.

\textbf{In-domain evaluation in LIBERO.}
As shown in Table~\ref{tab:libero_results}, IDR improves all four VLA backbones in LIBERO.
The largest gains are observed on VLA-Adapter ($+1.40$) and $\pi_{0.5}$ ($+1.25$), where our test-time adaptation provides the most benefit.
The smaller gain on X-VLA ($+0.45$) reflects its already proprioception-dominant baseline, which leaves less room for visual-guided correction.

\begin{figure*}[t]
\IEEEtopfloatpad
    \centering
    \includegraphics[width=\textwidth]{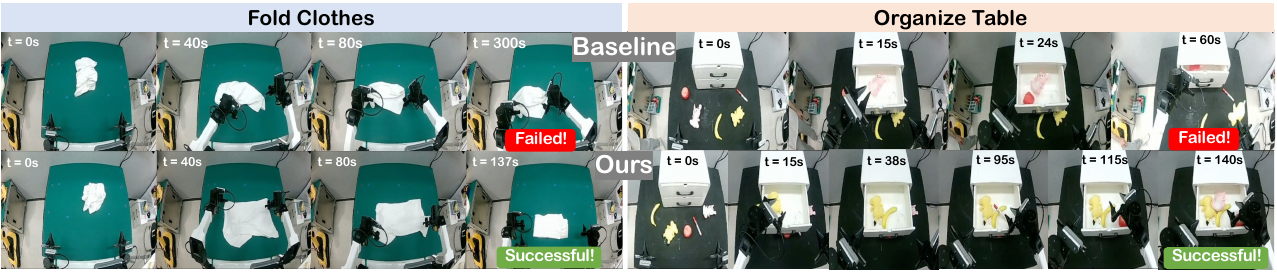}
    \caption{
    Real-world qualitative comparisons on Fold Clothes and Organize Table.
    The baseline model stalls before task completion, leading to failed executions. In contrast, IDR generates more task-directed actions and successfully completes both manipulation sequences via test-time adaptation. More visualizations are provided in multimedia materials.
    }
    \label{fig:real_env}
    \vspace{-10pt}
\end{figure*}

\textbf{Out-of-domain evaluation on SIMPLER and CALVIN.}
To evaluate robustness under distribution shifts, we evaluate IDR on SIMPLER and CALVIN. Specifically, SIMPLER evaluates Google Robot and WidowX under visual variation in backgrounds, lighting, distractors, and table textures, while CALVIN adopts the protocol where models trained on environments A-C are evaluated in the unseen environment D.
As shown in Tables~\ref{tab:simpler} and~\ref{tab:calvin}, IDR improves average performance across diverse environments, embodiments, and task horizons. On SIMPLER, the largest gain is achieved in the Variant Aggregation (VA) setting ($+6.33$), where visual adaptability is particularly critical. On CALVIN, IDR improves both the average completed length (from $4.22$ to $4.44$) and the $5/5$ completion rate (from $70.5\%$ to $77.3\%$), demonstrating its effectiveness in long-horizon execution.

Overall, IDR achieves considerable gains in both in-domain and out-of-domain evaluations, demonstrating that our test-time adaptation improves frozen VLA models under distribution shifts from visual and environmental variation.

\begin{table}[t]
    \centering
    \caption{SIMPLER benchmark results on X-VLA (\%). VM denotes Visual Matching and VA denotes Variant Aggregation.}
    \label{tab:simpler}
    \scriptsize
    \setlength{\tabcolsep}{4pt}
    \resizebox{\columnwidth}{!}{%
    \begin{tabular}{llccccc}
    \toprule
    Env. & Method & Coke Can & Move Near & Open Close & Place In & Avg \\
    \midrule
    Google VM & X-VLA & \textbf{98.33} & \textbf{97.08} & 71.76 & 57.41 & 81.15 \\
    \rowcolor{gray!15} Google VM & +IDR {(+0.98)} & 96.67 & 96.67 & \textbf{75.92} & \textbf{59.26} & \textbf{82.13} \\
    \midrule
    Google VA & X-VLA & 86.25 & \textbf{83.83} & 60.95 & 59.26 & 72.57 \\
    \rowcolor{gray!15} Google VA & +IDR {(+6.33)} & \textbf{89.63} & 83.80 & \textbf{68.10} & \textbf{74.07} & \textbf{78.90} \\
    \midrule
    \multicolumn{2}{l}{} & Blocks & Eggplant & Spoon & Place & Avg \\
    \midrule
    WidowX & X-VLA & 75.00 & \textbf{100.00} & \textbf{100.00} & \textbf{100.00} & 93.75 \\
    \rowcolor{gray!15} WidowX & +IDR {(+2.08)} & \textbf{87.50} & \textbf{100.00} & \textbf{100.00} & 95.80 & \textbf{95.83} \\
    \bottomrule
    \end{tabular}
    }
    \vspace{-10pt}
\end{table}

\begin{table}[t]
    \centering
    \caption{CALVIN ($ABC \rightarrow D$) results. Avg Len is the average number of completed subtasks.}
    \label{tab:calvin}
    \footnotesize
    \setlength{\tabcolsep}{3pt}
    \begin{tabularx}{\columnwidth}{@{}lYYYYYY@{}}
    \toprule
    Method & 1/5 & 2/5 & 3/5 & 4/5 & 5/5 & Avg~Len \\
    \midrule
    X-VLA & 95.7 & 90.9 & 86.0 & 78.8 & 70.5 & 4.22 \\
    \rowcolor{gray!15}
    +IDR {(+0.22 len)} & \textbf{98.1} & \textbf{94.2} & \textbf{89.9} & \textbf{84.3} & \textbf{77.3} & \textbf{4.44} \\
    \midrule
    VLA-Adapter & 97.1 & 93.8 & 88.8 & 82.5 & 76.4 & 4.39 \\
    \rowcolor{gray!15}
    +IDR {(+0.04 len)} & \textbf{97.7} & \textbf{94.1} & \textbf{89.3} & \textbf{83.8} & \textbf{77.8} & \textbf{4.43} \\
    \bottomrule
    \end{tabularx}
    \vspace{-10pt}
\end{table}

\begin{figure}[t]
    \centering
    \includegraphics[width=0.45\textwidth]{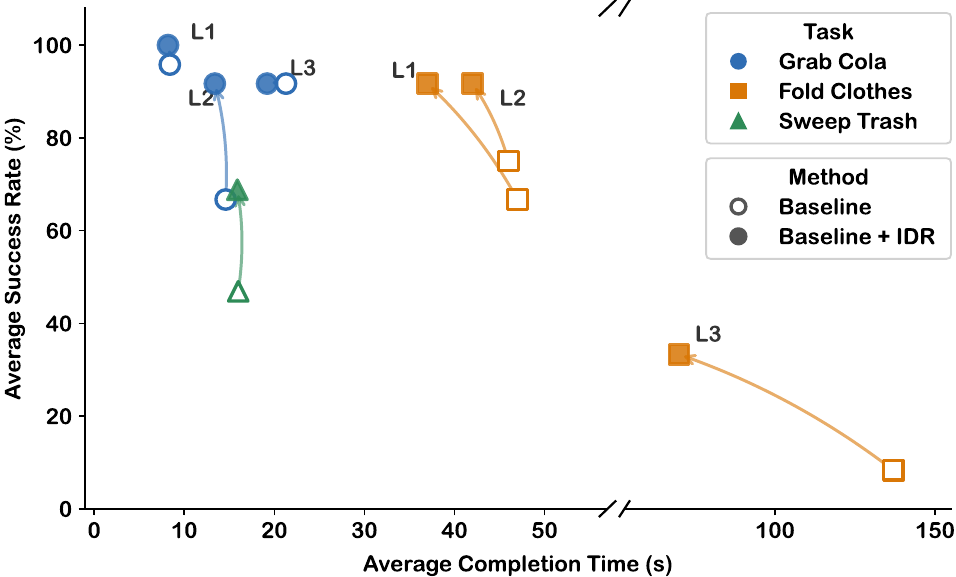}
    \caption{Real-world efficiency--success trade-off. Success rate and average completion time are reported for each task. Upward-leftward arrows (Baseline $\rightarrow$ IDR) indicate simultaneous improvements in both.}
    \label{fig:realworld_panels}
    \vspace{-10pt}
\end{figure}

\begin{table}[t]
    \centering
    \caption{Organize Table: long-horizon sequential task results. Avg Len is the average number of completed subtasks.}
    \label{tab:organize_table}
    \footnotesize
    \setlength{\tabcolsep}{3pt}
    \begin{tabularx}{\columnwidth}{@{}lYYYYYY@{}}
    \toprule
    Method & 1/5 & 2/5 & 3/5 & 4/5 & 5/5 & Avg~Len \\
    \midrule
    Baseline & 83.3 & 66.7 & 33.3 & 33.3 & 0.0 & 2.17 \\
    \rowcolor{gray!15}
    +IDR {(+1.83 len)} & \textbf{100.0} & \textbf{100.0} & \textbf{100.0} & \textbf{66.7} & \textbf{33.3} & \textbf{4.00} \\
    \bottomrule
    \end{tabularx}
    \vspace{-15pt}  
\end{table}

\subsection{Real-World Experiments}

\textbf{Experimental setup.} We evaluate IDR on real-world tasks using a dual-arm ARX5 platform, as shown in Fig.~\ref{fig:real_env}. We build our VLA model upon Qwen3-VL-4B~\cite{bai2025qwen3vl} with a standard Transformer action head, providing a clean baseline while minimizing potential confounding factors introduced by other modules. We train the model on approximately 130 hours of robot demonstration
data while keeping the visual encoder frozen.
The evaluation covers four task categories with different manipulation challenges.
\textbf{Grab Cola} evaluates grasping, transfer, and bimanual coordination across three difficulty levels (L1–L3), ranging from direct pick-and-place to cross-arm transfer and placement (12 trials per level).
\textbf{Fold Clothes} evaluates deformable object manipulation using clothes of four colors across three levels of deformation and folding complexity (L1–L3; 12 trials per level).
\textbf{Sweep Trash} evaluates contact-rich sweeping of scattered objects into a dustpan (16 trials total).
\textbf{Organize Table} evaluates long-horizon sequential manipulation, requiring five objects to be placed into a drawer in a specific order (6 trials total).

Fig.~\ref{fig:realworld_panels} summarizes the results across all tasks except Organize Table, whose subtask-wise completion rates are reported in Table~\ref{tab:organize_table}. Overall, IDR improves the average success rate from $56.5\%$ to $75.3\%$ ($+18.8\%$) while reducing the average completion time among successful trials from $53.6$s to $41.5$s. The gains are particularly pronounced on tasks requiring fine-grained visual guidance and contact-rich manipulation, such as Fold Clothes L3 ($+25.0\%$), Grab Cola L2 ($+25.0\%$), and Sweep Trash ($+21.9\%$). On the challenging 5-step Organize Table task, IDR achieves a $33.3\%$ full-completion rate compared with $0\%$ for the baseline, while increasing the average completed steps from $2.17$ to $4.00$. These results demonstrate that our test-time adaptation improves both manipulation accuracy and long-horizon execution in real-world settings.

\textbf{Inference efficiency.}
IDR requires three forward passes per step, which introduces additional computational cost.
In simulation, this overhead has no impact on task completion because environments advance on discrete step boundaries.
In real-world deployment, however, the overall completion time depends not only on inference latency but also on execution efficiency.
Despite the additional forward passes, IDR produces more task-directed actions and reduces unnecessary motions, as illustrated by comparative cases in Fig.~\ref{fig:real_env}.
As a result, we observe a lower average completion time (from $53.6$s to $41.5$s) in our real-world experiments, indicating that improved action quality can partially compensate for the added inference cost in execution.

\subsection{Ablation Studies}

\begin{table}[t]
    \centering
    \caption{Ablation of counterfactual intervention strategies on
    $\pi_{0.5}$ in LIBERO. Gaussian noise adds 50\% std of the input image, while mean-value padding uses the mean value.}
    \label{tab:intervention}
    \footnotesize
    \setlength{\tabcolsep}{3pt}
    \resizebox{\columnwidth}{!}{
    \begin{tabular}{lccccc}
    \toprule
    Strategy & Spatial & Object & Goal & Long & Avg \\
    \midrule
    Baseline & 97.50 & 95.00 & 97.00 & \textbf{95.50} & 96.25 \\
    +IDR (Gaussian noise) & 97.80 & 98.20 & \textbf{98.00} & 92.00 & 96.50 \\
    +IDR (mean-value padding) & 96.80 & 99.20 & 97.00 & 92.80 & 96.45 \\
    +IDR (zero-padding) & \textbf{98.00} & \textbf{100.00} & 97.00 & 95.00 & \textbf{97.50} \\
    \bottomrule
    \end{tabular}
    }
    \vspace{-10pt}
\end{table}

\textbf{Intervention strategies.}
We further evaluate alternative counterfactual interventions, as shown in Table~\ref{tab:intervention}.
Both Gaussian noise and mean-value padding improve the baseline,
demonstrating that IDR is compatible with different intervention
strategies. Zero-padding achieves the best result, improving the average
success rate from 96.25\% to 97.50\%. We hypothesize that zero-padding
provides a deterministic and more complete removal of modality information, yielding a cleaner counterfactual signal, whereas others may retain partial information or introduce additional input artifacts.

\begin{figure}[t]
\IEEEtopfloatpad
    \centering
    \includegraphics[width=0.45\textwidth]{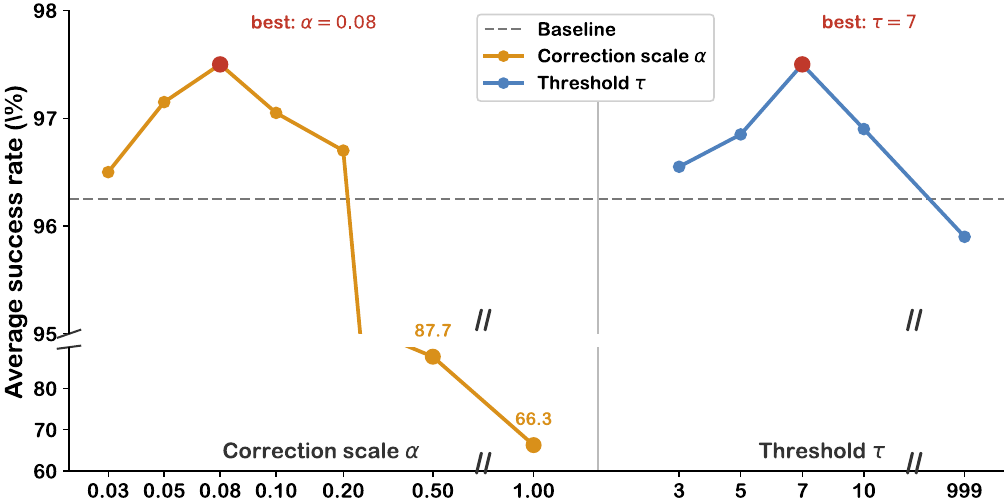}
    \caption{Hyperparameter ablation on $\pi_{0.5}$ in LIBERO.
    Moderate correction scales and intervention thresholds yield the best performance, while excessive correction or uniform intervention degrades action generation.
    }
    \label{fig:ablations_hypers}
    \vspace{-10pt}
\end{figure}

\textbf{Correction scale (\(\alpha\)) and intervention threshold (\(\tau\)).}
Fig.~\ref{fig:ablations_hypers} analyzes the two primary hyperparameters.
For the correction scale $\alpha$, performance peaks at $0.08$, while nearby values such as $0.05$ and $0.10$ produce comparable results.
Large values ($\alpha \geq 0.5$) degrade performance, indicating that excessive correction can destabilize action generation.
Overall, the model remains robust within a moderate range of $\alpha$.
For the intervention threshold, small values rarely activate the gate, and therefore provide limited improvement.
In contrast, uniform intervention across all timesteps ($\tau=999$) performs below the baseline.
The best performance is obtained at $\tau=7$ (approximate mean value), where refinement is applied only to predictions with low diagnosed visual importance.
This supports the need for selective intervention rather than applying correction uniformly throughout execution.

\textbf{Modules of action refiner.}
We further evaluate each component of the causality-aware action refiner, as shown in Table~\ref{tab:module_ablation}.
The full model (Mode~E) achieves the best success rate of $97.50\%$, improving over the baseline by $+1.25\%$.
Removing the selective gate (Mode~D) or the proprioceptive regularizer (Mode~A) leads to degradation compared with the full model.
This indicates that both selective refinement and bounded proprioceptive regularization are important for stable refinement.
Removing the clip bound (Mode~B) or adaptive proprioceptive weighting (Mode~C) also reduces performance, confirming that these components help regularize the residual update.
Additionally, reversing the intervention direction (Mode~F) drops performance below the baseline.
This shows that the improvement comes not merely from adding interventions, but from steering the action along the diagnosed causal effect direction.
Fig.~\ref{fig:phase_aligned_effect_ratio} further illustrates the phase-dependent behavior of the refinement in LIBERO.
Mode~E changes the visual importance ratio around execution phase transitions, whereas Mode~F fails to produce the intended adjustment, confirming the model's sensitivity to both the magnitude and direction of causal intervention.

\begin{table}[t]
    \centering
    \caption{Module ablation on $\pi_{0.5}$ in LIBERO.}
    \label{tab:module_ablation}
    \footnotesize
    \setlength{\tabcolsep}{3pt}
    \begin{tabularx}{\columnwidth}{@{}lCCCCC@{}}
    \toprule
    Mode & Spatial & Object & Goal & Long & Avg \\
    \midrule
    Baseline & 97.50 & 95.00 & 97.00 & 95.50 & 96.25 \\
    A (w/o proprioception) & 97.60 & 98.00 & 96.20 & 93.40 & 96.30 \\
    B (w/o clip) & 97.80 & 98.40 & 97.00 & 93.40 & 96.65 \\
    C (w/o adaptive $w_{\text{prop}}$) & 97.80 & 99.20 & \textbf{97.20} & 93.60 & 96.95 \\
    D (w/o gate) & 97.60 & 98.00 & 96.20 & 93.00 & 96.20 \\
    \rowcolor{gray!15}
    E (ours) & \textbf{98.00} & \textbf{100.00} & 97.00 & \textbf{95.00} & \textbf{97.50} \\
    F (negated $\alpha$, $\beta$) & 98.40 & 96.80 & 96.80 & 88.20 & 95.05 \\ 
    \bottomrule
    \vspace{-10pt}
    \end{tabularx}
\end{table}

\begin{table}[ht]
    \centering
    \caption{Proprioception-guided refinement on X-VLA in LIBERO.}
    \label{tab:xvla_symmetric}
    \footnotesize
    \setlength{\tabcolsep}{3pt}
    \begin{tabularx}{\columnwidth}{@{}lCCCCC@{}}
    \toprule
    Method & Spatial & Object & Goal & Long & Avg \\
    \midrule
    X-VLA (baseline) & 92.60 & 99.20 & 98.00 & \textbf{94.40} & 96.05 \\
    \rowcolor{gray!15}
    +IDR (visual-guided) & \textbf{97.00} & \textbf{100.00} & 95.00 & 94.00 & 96.50 \\
    +IDR (proprioception-guided) & 92.20 & 50.60 & \textbf{98.20} & 68.40 & 77.35 \\
    \bottomrule
    \end{tabularx}
    \vspace{-10pt}
\end{table}

\textbf{Which modality should guide the refinement?}
We next examine whether the refinement signal should follow the dominant modality of a given backbone.
Since Observation~2 identifies X-VLA as proprioception-dominant in LIBERO, we construct a proprioception-guided variant that uses $\Delta_{\text{prop}}$ as the main correction signal, with corresponding gate and weight inversions while keeping other settings unchanged.

Table~\ref{tab:xvla_symmetric} reports the results.
The proprioception-guided variant drops to $77.35\%$ average success, far below both the X-VLA baseline ($96.05\%$) and visual-guided IDR ($96.50\%$).
The degradation is especially severe on the Object suite, decreasing from $99.20\%$ to $50.60\%$.
This reveals an asymmetry between visual and proprioceptive effects.
Even for a proprioception-dominant model, proprioceptive effects are not suitable as the main refinement signal.
We hypothesize that while vision captures external scene semantics for task-level correction, proprioception is tightly coupled with low-level control and thus may disrupt execution stability.

\section{Conclusion}
\label{sec:conclusion}

We have presented IDR, an infer-diagnose-refine framework for training-free modality adaptation in VLA models. 
IDR diagnoses the causal effects of visual observations by comparing factual and counterfactual predictions, and uses the causal effects to refine the action predictions at test time.
Our analysis reveals that visual importance is not fixed, but varies across VLA architectures, environments, and execution phases. 
Experiments on simulation benchmarks and real-world manipulation tasks show reliable improvements across multiple VLA backbones, demonstrating the effectiveness of test-time causal inference for action refinement.
The primary limitation is the requirement of three forward passes per control step, which increases inference latency.
Future work will explore distilling these online causal diagnostics into the model during training, enabling implicit modality-aware adaptation without test-time computational overhead.

\end{document}